\documentclass[10pt,english]{article}

\usepackage{microtype}
\usepackage{graphicx}
\usepackage{subfigure}
\usepackage{booktabs} 
\usepackage{multirow} 
\usepackage{array}    
\usepackage{caption}  
\usepackage{algorithm}  
\usepackage{algorithmicx} 
\usepackage{algpseudocode} 
\usepackage{pifont}
\usepackage{stfloats}  

\usepackage{amsmath}  
\usepackage{amssymb}
\usepackage{mathtools}
\usepackage{amsthm}
\usepackage{tikz}
\usepackage[most]{tcolorbox}

\usepackage{hyperref}

\usepackage[capitalize,noabbrev]{cleveref}

\usepackage{fullpage}
\usepackage{amsmath,amssymb,amsthm,graphicx,subfigure,hyperref,algorithm,algpseudocode,bbm,mathtools,caption,multirow}
\usepackage[noadjust]{cite}
\hypersetup{colorlinks=true, citecolor=blue, urlcolor=blue}
\usepackage{parskip}

\usepackage{algorithm}
\usepackage{algpseudocode}

\theoremstyle{plain}

\title{\textbf{CuDIP: Enhancing Theorem Proving in LLMs via Curriculum Learning-based Direct Preference Optimization}}

\author{
    \textbf{Shuming Shi\textsuperscript{1}} \quad 
    \textbf{Ruobing Zuo\textsuperscript{1}} \quad 
    \textbf{Gaolei He\textsuperscript{2}} \quad 
    \textbf{Jianlin Wang\textsuperscript{2}} \quad \\
    \textbf{Chenyang Xu\textsuperscript{1}} \quad 
    \textbf{Zhengfeng Yang\textsuperscript{1}} \quad 
    \\[0.5em]
    {\textbf{\textsuperscript{1}East China Normal University,}}
    {\textbf{\textsuperscript{2}Henan University}}
    \\[0.5em]
    {\{smshi, rbzuo\}@stu.ecnu.edu.cn},
    {\{hegaolei, jlwang\}@henu.edu.cn},\\
    {\{cyxu, zfyang\}@sei.ecnu.edu.cn} \quad
}

\date{}
\begin{document}
\maketitle







\begin{abstract}
Automated theorem proving (ATP) is one of the most challenging mathematical reasoning tasks for Large Language Models (LLMs). 
Most existing LLM-based ATP methods rely on supervised fine-tuning, 
which results in a limited alignment between the theorem proving process and human preferences. 
Direct Preference Optimization (DPO), which aligns LLMs with human preferences, has shown positive effects for certain tasks. 
However, the lack of high-quality preference data for theorem proving presents a significant challenge. 
In this paper, we innovatively apply DPO to formal automated theorem proving and introduces a Curriculum Learning-based DPO Iterative Theorem Proving (CuDIP) method. 
Specifically, we propose a method for constructing preference data which utilizes LLMs and existing theorem proving data to enhance the diversity of the preference data 
while reducing the reliance on human preference annotations. 
We then integrate this preference data construction method with curriculum learning to iteratively fine-tune the theorem proving model through DPO. 
Experimental results on the MiniF2F and ProofNet datasets demonstrate the effectiveness of the proposed method.


\end{abstract}

\section{Introduction}
\label{Instruction}
Despite the success of LLMs in various fields 
\cite{ziegler2019fine,lightman2023let,wang2024mathshepherd,guo2024deepseekCODER},  
they still exhibit significant limitations in mathematical reasoning tasks, particularly in theorem proving. 
Enhancing the performance of LLMs in these tasks remains a challenging problem, especially for automated theorem proving in strictly formalized interactive theorem provers (ITPs) such as Lean \cite{lean2015lean} and Isabelle \cite{paulson1990isabelle}. Many studies integrate LLMs with ITPs for automated theorem proving. Existing approaches generally fall into two categories: using LLMs to provide stepwise proof strategies combined with search algorithms to complete proofs \cite{polu2020gpt-f,han2021PACT,lample2022hypertree}, 
and leveraging LLMs to independently or informally generate complete proofs \cite{jiang2022dsp,wang2023lego}. 
While these methods have achieved some success, most rely on supervised fine-tuning, resulting in limited alignment with human preferences, 
and the effectiveness of theorem proving remains constrained.

In recent years, LLM fine-tuning methods aligned with human preferences have gained significant attention. 
Recent studies demonstrate that preference-based optimization for LLM fine-tuning can yield substantial benefits \cite{ziegler2019fine,stiennon2020learning}. 
Reinforcement Learning with Human Feedback (RLHF) \cite{christiano2017RLHF} is a well-established and effective method for aligning models with human preferences, typically consisting of two steps: (1) training a reward model, and (2) optimizing it using reinforcement learning methods such as PPO \cite{schulman2017PPO}. 
However, this approach has two primary limitations: 1) it is relatively complex; and 2) it entails significant computational overhead.

To address the aforementioned issue, 
Direct Preference Optimization (DPO) \cite{dpo2024} offers a simpler and more efficient alternative. 
DPO eliminates the need for training a reward model, directly optimizing from preferences, which makes it both straightforward and effective. 
DPO has shown promising results in certain mathematical reasoning tasks \cite{lu2024step,pal2024smaug}, 
but has yet to be applied to theorem proving tasks. 
However, a key challenge for preference-based optimization methods in LLM fine-tuning is the requirement for high-quality human preference labels, which are difficult to obtain.  

To overcome the challenges mentioned above, 
this paper proposes a method for automated theorem proving that combines DPO and curriculum learning, named \textbf{C}urriculum {L}earning-based \textbf{D}PO \textbf{I}terative Theorem \textbf{P}roving (CuDIP). 
To the best of our knowledge, this is the first work to integrate DPO with automated theorem proving, filling a gap in DPO-based automated theorem proving.
To address the challenge of insufficient theorem proving preference data, this paper proposes a method for constructing preference data based on fine-grained scoring by LLMs, 
which reduces the reliance on human annotations by utilizing LLMs for preference data construction, and enhances the diversity of positive samples in the generated preference data through the incorporation of fine-grained preference scoring.
Using the constructed preference data in conjunction with curriculum learning, we propose a process for iteratively training a prover based on DPO, which enhances the LLM's ability to solve difficult problems through progressively challenging DPO iterations. 
Specifically, we 
categorizing automated theorem proving problems into curriculum data based on predefined difficulty levels, and iteratively construct and fine-tune the DPO preference data following the principles of curriculum learning.
Experimental results indicate that the pass rate reaches 38.5\% on the MiniF2F dataset and 12.7\% on the ProofNet dataset, 
surpassing the baseline methods and highlighting the superior performance of the proposed method. 

Our contributions can be summarized as follows:
\begin{itemize}
    \item We propose a method for constructing DPO preference data based on fine-grained preference scoring by Large Language Models (LLMs), 
    which facilitates the generation of diverse preference data while reducing reliance on human annotations.
    \item Based on the proposed preference data construction method and curriculum learning, we introduce Curriculum Learning-based DPO Iterative Theorem Proving (CuDIP) method, which enhances the capability of LLMs in autumated theorem proving.
    \item Experimental results 
    demonstrate that the proposed method outperforms all baseline approaches, achieving a maximum improvement of 7.4\% on the MiniF2F-valid benchmark, thereby effectively enhancing the theorem proving capabilities of LLMs.
\end{itemize}


%

 



\section{Related Works} 
\textbf{Automated Theorem Proving with LLMs.}
The rapid advancement of Large Language Models (LLMs) 
has spurred significant progress in automated theorem proving.
Various approaches integrate interactive theorem provers (ITP) such as 
Lean \cite{de2015lean}, Isabelle \cite{paulson1990isabelle}, and Metamath \cite{megill2019metamath} with language models. 
A prominent method, exemplified by GPT-f \cite{polu2020gpt-f}, involves the language model generating the next proof step based on the current proof state, followed by tree search to find a complete proof for the theorem. 
PACT \cite{han2021PACT} jointly trains a language model with a strategy prediction objective for interactive theorem proving, where the auxiliary task for extracting self-supervised signals is used. 
HTPS \cite{lample2022hypertree} improves the automated theorem proving process by enhancing the MCTS-based proof search strategy. 
Another approach employs LLMs to derive the complete proof of a theorem, either directly or with the assistance of informal proof languages \cite{jiang2022dsp,lin2024leanstar,wang2023lego}. 
However, most approaches rely on supervised fine-tuning (SFT), which may not align well with human preferences, potentially limiting their proving capabilities.



\textbf{Aligning LLMs with Human Preference.} 
Reinforcement Learning from Human Feedback (RLHF) \cite{christiano2017RLHF} is a classical and effective method for aligning LLMs with human preferences. This approach involves initially training a reward model, followed by optimization using reinforcement learning algorithms such as 
PPO \cite{schulman2017PPO}, 
leading to substantial success in models like 
ChatGPT 
\cite{radford2019lGPT,brown2020GPT,achiam2023gpt}, 
LLaMA \cite{touvron2023llama,dubey2024llama}, 
and Claude \cite{bai2022training}. 
However, using PPO for RLHF is a complex and computationally expensive process. Motivated by this, Direct Preference Optimization (DPO) \cite{dpo2024} has emerged as an effective alternative. 
DPO directly optimizes using preference data, without the need to train a reward model. 
Several works \cite{ethayarajh2024kto,hong2024orpo,lai2024stepdpo,xu2024contrastive,zeng2024token,azar2024dpo_variant}  
have proposed variations based on DPO. 
The simplicity and efficiency of DPO have enabled its application in various downstream tasks, such as 
mathematical reasoning \cite{xu2024chatglm,jiao2024learning,lu2024step-controlled-dpo}, 
multimodal tasks \cite{majumder2024tango}, 
and 
summarization \cite{dpo2024}.  
However, DPO still relies on high-quality preference labels to generate preference data, which requires costly manual annotation. 
In this paper, we propose a method for constructing DPO preference data and 
the generated preference data is utilized in DPO iterative training to enhance the theorem proving capabilities of LLMs.


\textbf{Curriculum Learning.} 
Curriculum learning \cite{bengio2009curriculum} 
is a method that simulates the human learning process by progressing from simpler to more complex tasks. 
Chang et al. \cite{chang2021does} applied curriculum learning to data-to-text generation, achieving improvements in generation quality. 
By strategically organizing the learning trajectory, curriculum learning enhances model performance or training efficiency and has shown promise in mathematical reasoning tasks such as theorem proving. 
Polu et al. \cite{polu2022formalCL} introduced a curriculum of progressively harder statements by synthesizing inequalities with increasing difficulty, 
while LeanAgent \cite{kumarappan2024leanagent} categorized theorems into three levels of complexity and utilized curriculum learning to train on mathematical repositories. 
In this paper, we propose a curriculum learning iterative training method based on the difficulty of theorem proof data and demonstrate its effectiveness.


\section{Preliminaries and Notation}
\subsection{Preliminaries}
\label{section_3.1}
\textbf{Theorem Proving in Lean.} 
Lean \cite{lean2015lean,moura2021lean4} is a reliable interactive theorem proving assistant. Unlike natural language-based theorem proving, Lean requires a more rigorous proof process. Specifically, the theorem proving in Lean is interactive, where each step is carefully checked. 
In Lean, the theorem statement serves as the initial goal, and each formalized proof strategy, called a "tactic", is applied to reach new subgoals. 
The proof is considered complete when the "no goals" state is reached.

\textbf{Direct Preference Optimization (DPO).} DPO \cite{dpo2024} is an offline reinforcement learning method that aligns the outputs of Large Language Model (LLM) with human preferences. 
Unlike Reinforcement Learning from Human Feedback (RLHF) \cite{christiano2017RLHF}, which first trains a reward model and then applies reinforcement learning, DPO directly optimizes based on preference data. Each sample in the DPO preference dataset is a triplet $\left(x, y_w, y_l\right)$, where $y_w$ is the preferred response and $y_l$ is the dispreferred response, both corresponding to the input prompt $x$. The training objective is to minimize the following loss:

\label{eq_loss}
\begin{equation}
\mathcal{L}_{DPO}({\pi}_\theta;{\pi}_{\text{ref}}) = - \mathbb{E}_{(x, y_w, y_l) \sim \mathcal{D}} \Big[ \log \sigma \Big( \beta \log \frac{\pi_\theta(y_w | x)}{ \pi _{\text{ref}}(y_w | x)} 
- \beta \log \frac{\pi_\theta(y_l | x)}{\pi_{\text{ref}}(y_l | x)} \Big) \Big],
\end{equation}

where $\mathcal{D}$ is the preference dataset, $\sigma$ is the sigmoid function, $\pi_{\theta}$ is the model to be optimized, initialized as $\pi_{\text{ref}}$, and $\beta$ is a hyperparameter used to control the degree of divergence between $\pi_{\theta}$ and $\pi_{\text{ref}}$.


\subsection{Problem Definition}  
\label{Section:3.2}
This paper investigates the problem of automated theorem proving in the Lean language. The theorem proving process can be abstracted as a tree search process, where the root node represents the theorem statement to be proven. 
Each node in the tree corresponds to a proof state $s$ 
(i.e., a proof subgoal and corresponding premises), 
and each edge represents a proof tactic $t$. 
The initial state $s_0$ represents the target theorem statement to be proved. Finding a valid proof for $s_0$ entails identifying a complete path 
starting from $s_0$ that successfully resolves the proof goals along the path.

\subsection{Notation}  
In this paper, we define $\mathcal{D}$ as the existing formalized theorem dataset with complete proof steps, 
and $\mathcal{C}= \{\mathcal{C}_1,\mathcal{C}_2,...,\mathcal{C}_n\}$ as the curriculum dataset derived from $\mathcal{D}$.  
$\mathcal{P}$ represents the prover model, where $\mathcal{P}_{n}$ denotes the prover model at the $n$-th iteration, and $\mathcal{P}_{0}$ refers to the base prover obtained through supervised fine-tuning (SFT) of the base model. 
$\mathcal{G}$ represents the score generator model used to produce preference data, 
and $\mathcal{G}_{n}$ denotes the score generator model at the $n$-th iteration.

The theorem proving state at step $i$ is denoted as $s_i$. 
$d_i$ denotes the \textit{distance} from $s_i$ to completing the proof, and $DIF_i$ represents the \textit{difficulty} of $s_i$. 
$\mathcal{T}$ represents the set of all candidate tactics for a given proof goal. 
$\mathcal{SC}$ denotes the dataset of corresponding preference scores for the candidate tactics in $\mathcal{T}$. 
$\mathcal{D}_P$ represents the preference dataset for DPO.

%


\section{Methodology}   

\begin{figure*}[t]
\centering
\includegraphics[width=0.9\linewidth]{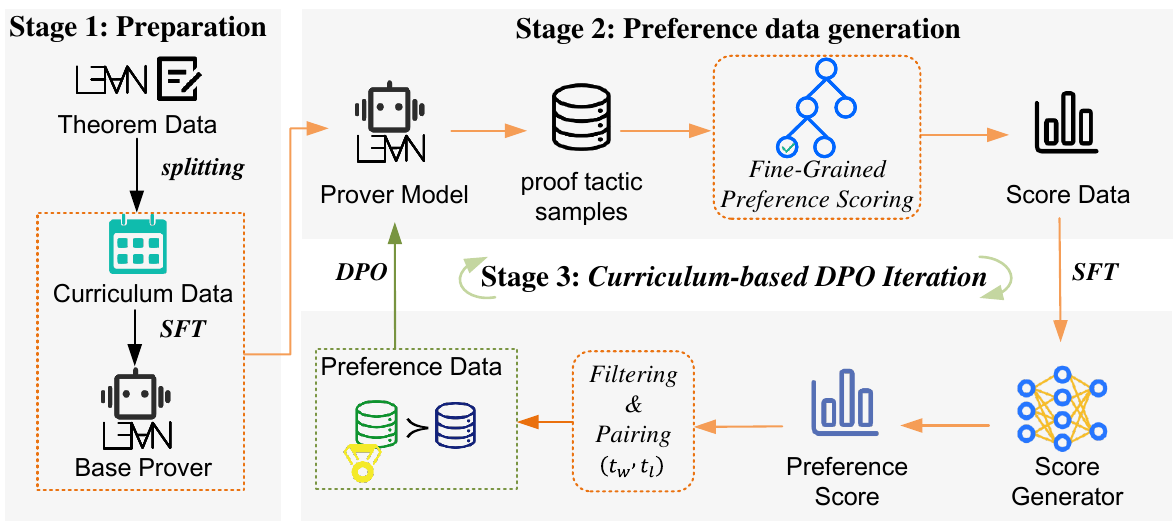}
\caption{Overview of \textbf{C}urriculum {L}earning-based \textbf{D}PO \textbf{I}terative Theorem \textbf{P}roving  (\textbf{CuDIP}). 
\textbf{Stage 1: Preparation.} 
The original data is partitioned to create curriculum data, and the base prover is trained using Supervised Fine-Tuning (SFT).
\textbf{Stage 2: Preference Data Generation.} 
A score generator is trained to assign fine-grained scores to candidate strategies for a given proof state, followed by filtering and pairing to generate preference data. 
\textbf{Stage 3: Curriculum-based DPO Iteration.} 
The prover model undergoes iterative DPO fine-tuning based on the generated preference dataset, refining both the preference data and the prover model through successive iterations.
}
\label{overview_1}
\end{figure*}
In this section, we introduce the proposed \textbf{C}urriculum {L}earning-based \textbf{D}PO \textbf{I}terative Theorem \textbf{P}roving (\textbf{CuDIP}). 
First, we present our key insights and an overview of the framework of the proposed CuDIP method, which consists of three stages: 
(1) Preparation, 
(2) Preference Data Generation, 
and (3) Curriculum-based DPO Iteration. 
Subsequently, we provide a detailed introduction to each of these stages. 

\subsection{Key Insights and Framework Overview}

Our proposed approach is based on the following observations and assumptions.
\begin{enumerate}

\item The preference-based LLM fine-tuning method DPO can be applied to theorem-proving tasks in LLMs to enhance reasoning capabilities and align better with human preferences. 

\item Similar to the human learning process, where knowledge is acquired progressively from simple to complex, training LLMs in a stepwise manner—from easy to difficult—may enhance their learning process and improve their ability to tackle more complex problems.


\item The off-the-shelf expert data from the theorem proving process can serve as positive examples for DPO training. Additionally, existing language models can generate counterexamples and more diverse DPO training data based on the available theorem-proving data.
\item When constructing DPO preference data, applying more fine-grained preference scoring to the data prior to forming preference pairs enhances the diversity of positive examples in the preference data.  
\end{enumerate}

Based on the aforementioned key insights, we propose 
a curriculum learning-based DPO iterative theorem proving method (CuDIP) 
for the theorem proving problem defined in Section \ref{Section:3.2}. 
To address the lack of preference data in theorem proving, 
we propose a fine-grained scoring-based preference data generation method, which utilizes existing theorem proving data and LLMs to generate more diverse preference data, thereby reducing the reliance on human annotations. 
We then apply curriculum learning-based DPO iteration to the prover, leveraging the proposed preference data generation method and curriculum learning.

Figure \ref{overview_1} gives an overview of our proposed CuDIP method. 
Specifically, the overall algorithm can be divided into the following three stages: 

\textbf{Stage 1: Preparation.} 
The existing formal theorem proving dataset  $\mathcal{D}$ is partitioned based on the \textit{difficulty} of the tasks to construct a curriculum learning dataset $\mathcal{C}$. The base model of the prover is then fine-tuned in a supervised manner using the entire curriculum learning dataset $\mathcal{C}$, resulting in the base prover $\mathcal{P}_{0}$. 

\textbf{Stage 2: Preference Data Genertion.} (1)Based on the current prover model and a subset of 
the new round of curriculum learning data $\mathcal{C}_{n}$, the preference score data $\mathcal{SC}_{0}$ for training the score generator is derived through Fine-Grained Preference Scoring (FGPS) process. 
(2)Supervised fine-tuning (SFT) is applied to the current generator model using $\mathcal{SC}_{0}$, 
resulting in a new score generator model $\mathcal{G}_{n}$. 
(3)$\mathcal{G}_{n}$ is then used to score the candidate tactics across all states in $\mathcal{C}_{n}$, and the preference dataset $\mathcal{D}_P$ is obtained through \textit{filtering and pairing}.


\textbf{Stage 3: Curriculum-based DPO Iteration.}
The current prover model is fine-tuned using the generated preference data $\mathcal{D}_P$ through the DPO method. \textbf{Stage 2} and \textbf{3} are then repeated iteratively until no further performance improvement is observed. 

Next, we will provide a detailed technical description of the three stages of the proposed method. 


\subsection{Stage 1: Preparation}
\label{Stage1}
Before introducing the process of constructing curriculum data, we first define two concepts. 

\newtheorem{definition}{Definition}
\begin{definition}
\label{def:distance}
 (Distance). For a given proof tree $T$ of a theorem, the \textit{distance} of a proof state $s$ is defined as the number of proof tactics that need to be executed from $s$ 
 to the completion of the proof (i.e., reach the "no goals" state in Lean).  
\end{definition}

\begin{definition}
\label{def:difficulty}
(Difficulty). The \textit{difficulty} of resolving a proof goal $s$ is defined as the \textit{distance} in the proof tree from $s$ to the final proof goal.  
\end{definition}

\textbf{Curriculum Data Construction.} 
The curriculum dataset $\mathcal{C}$ is derived by partitioning the original theorem-proving dataset $\mathcal{D}$ based on  increasing \textit{difficulty} values. 
Specifically, for each theorem and its corresponding proof process in $\mathcal{D}$, we extract the proof state $s$ at each step of the proof process and the corresponding subsequent proof tactic $t$. 
We then calculate the \textit{difficulty} $DIF_i$ of each proof state $s_i$. 
According to definition 
\cref{def:difficulty}
, $DIF_i$ of $s_i$ is defined as the \textit{distance} $d_i$. 
This process results in intermediate data composed of triplets, $\mathcal{D}_{\text{triplet}}$, which can be formally expressed as:

\begin{equation}
\mathcal{D}_{triplet} = \{ (s_i, t_i, DIF_i) \mid i = 1, 2, \dots, n \},
\end{equation}
where $s_i$ represents the current theorem proving state, $t_i$ denotes the next proving tactic of $s_i$, and $DIF_i$ indicates the \textit{difficulty} of $s_i$. 

Then, divide the dataset $\mathcal{D}_{\text{triplet}}$ into subsets in ascending order of $DIF_i$. This results in a curriculum dataset $\mathcal{C}$ organized from easy to hard, which can be formally expressed as:

\begin{equation}
    \mathcal{C} = \{ \mathcal{C}_n \mid n \in \{DIF_i \mid (s_i, t_i, DIF_i) \in \mathcal{D}_{\text{triplet}} \} \},
\end{equation}
\begin{equation}
    \mathcal{C}_n = \{ (s_i, t_i, DIF_i) \mid DIF_i = n \},
\end{equation}  

where $\mathcal{C}$ is the set of all subsets $\mathcal{C}_n$, each corresponding to a unique \textit{difficulty} level $n$, ensuring that triplets with the same \textit{difficulty} are grouped together. In the subsequent training process, further processing is performed based on dataset $\mathcal{C}$. 

\textbf{Base Prover Training.} To achieve better proving performance, 
we first train the base model used as the prover model before the subsequent iterations. The entire curriculum learning dataset $\mathcal{C}$ is used to perform supervised fine-tuning (SFT) on the base model, resulting in the base prover $\mathcal{P}_{0}$.

After obtaining the curriculum learning data 
and the base prover, we next construct a preference dataset for DPO fine-tuning.

\subsection{Stage 2: Preference Data Generation}
\label{Stage2}
 We propose a method for constructing preference data by leveraging LLM and existing formal theorem-proving datasets.  
 First, we define a scoring rule and assign fine-grained preference scores to each theorem proof state and its corresponding candidate tactics in curriculum dataset acheived in \cref{Stage1}.
 Next, the data is filtered and paired based on the scores to form preference pairs, resulting in the final DPO preference dataset. To enhance the efficiency of the data generation process, we introduce a score generator during the preference scoring stage, replacing manually defined scoring rules. 
 The proposed method offers the following benefits: (1) reducing reliance on human annotations by utilizing LLMs for preference data construction, and (2) enhancing diversity of positive samples through fine-grained preference scoring. 
 The method proceeds as follows.


\textbf{Tactic Samples Generation.}  
For each proof state $s$ in the formalized course dataset $\mathcal{C}_n$, $k$ candidate tactics are generated using the current prover model $\mathcal{P}_{n-1}$, forming a candidate tactic set 
$\mathcal{T} = \{ t_{1},t_{2},...,t_{k} \} $ . 

\textbf{Fine-Grained Preference Scoring (FGPS).} 
For the candidate tactic set $\mathcal{T}$ originating from proof state $s$, all tactics in $\mathcal{T}$ are evaluated using the following scoring process.
\begin{enumerate}
    \item Execute each candidate tactic from $s$ to reach a new proof state and perform $n_{\text{attempt}}$ subsequent proof search attempts using the current prover $\mathcal{P}_{n-1}$ and Monte Carlo Tree Search (MCTS). 
    \item The score for a given candidate tactic $t$ of proof state $s$ is calculated using the following equation:
        \begin{equation}
            \text{Score} \left(t\mid s\right) = \frac{n_\text{success}}{n_\text{attempt}},
        \end{equation}
        where $n_\text{success}$ represents the number of successful proofs found after executing the proof tactic $t$, and $n_\text{attempt}$ is the number of subsequent proof search attempts for each candidate tactic.
\end{enumerate}

\begin{figure}[tbp]
\centering
\includegraphics[width=0.6\linewidth]{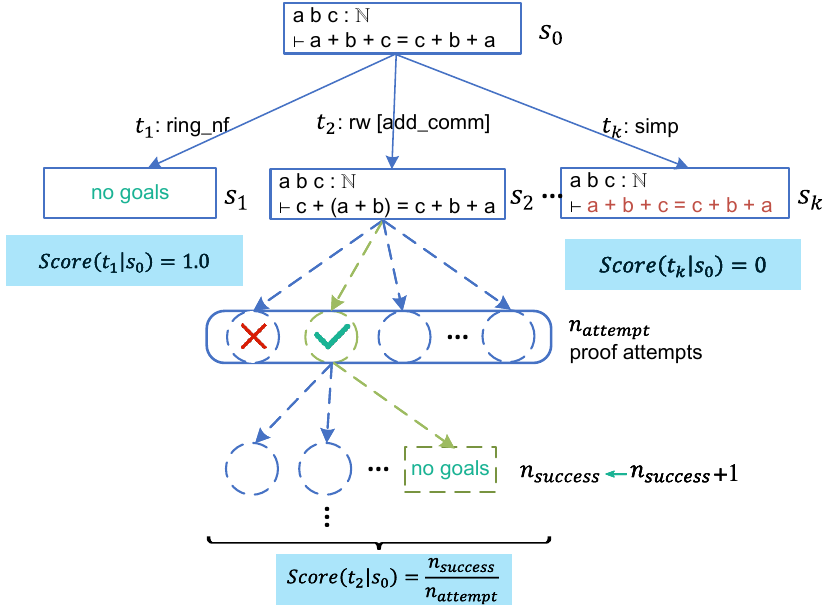}
\caption{
An example of fine-grained preference scoring of all candidate tactics for the initial state of the Lean theorem (a b c: $\mathbb{N}$): a + b + c = c + b + a.
}
\label{FGPS}
\end{figure}

As shown in \cref{FGPS}, an example of the proposed fine-grained scoring process is provided. 
However, The above scoring process necessitates multiple MCTS search attempts for each proof state, incurring significant time costs. Therefore, we randomly select a subset of 
the course data $\mathcal{C}_n$ for preference scoring to obtain the original score data $\mathcal{SC}^0$, and introduce a \textit{score generator} to score the remaining data in $\mathcal{C}_n$. 

\textit{REMARK}. For the remaining unscored data, language model-based scoring replaces the proof search scoring process, improving the efficiency of preference data construction.  

\textbf{Score Generator Training.} 
Using $\mathcal{SC}^0$, the existing language model $\mathcal{G}_{n-1}$ undergoes SFT to produce the score generator $\mathcal{G}_n$. The score generator is then used to evaluate the remaining data, resulting in the final scoring dataset $\mathcal{SC}$, which can be expressed as:
\begin{equation}
\mathcal{SC} = \{ (s_i, T_i) \mid i = 1, 2, \ldots, x \},
\end{equation}
\begin{equation}
T_i = \{ (t_{i}^{(j)}, \text{Score}\left(t_{i}^{(j)}\mid s_i\right)) \mid j = 1, 2, \ldots, k \}.
\end{equation}
Where \( s_i \) represents the \( i \)-th proof state in $\mathcal{C}_n$, 
\( T_i \) is the set of candidate proof tactics of $s_i$ and their corresponding preference scores, 
\( t_{i}^{(j)} \) denotes the \( j \)-th candidate tactic for \( s_i \), 
\( x \) is the total number of proof states in $\mathcal{C}_n$, and \( k \) is the number of candidate tactics for a given state. 




\textbf{Filtering and Paring.} 
Perform data filtering and pairing on the dataset $\mathcal{SC}$, which contains all states along with their corresponding candidate tactics and tactic preference scores.
Specifically, for a given proof state $s$, candidate tactics from its tactic set $\mathcal{T}$ are selected for pairwise comparison if the difference in their preference scores exceeds a certain threshold $\text{Th}$. 
A preference pair $\{s, t_w, t_l\}$ is considered valid if the following conditions are met:
\begin{equation}
    \text{Score}(t_w | s) > \text{Score}(t_l | s) ,
     \Delta_{Score_{t_w,t_l}} > \text{Th},
\end{equation}
where $\Delta_{Score_{t_w,t_l}}=\left| \text{Score}(t_w | s) - \text{Score}(t_l | s) \right|$.

After the filtering and pairing process, all the preference data $ \left(s, t_w, t_l \right)$  satisfying the conditions constitute the preference data set $\mathcal{D}_P$, which can be expressed as: 
\begin{equation}
    \mathcal{D}_P = \left\{ s^{(i)}, t_w^{(i)}, t_l^{(i)} \right\}_{i=1}^{N},
\end{equation}
where $N$ is the number of all qualified preference pairs in $\mathcal{C}_n$, and $s_i$ is the state corresponding to the preference candidate tactic pair.  




\subsection{Stage 3: Curriculum-based DPO Iteration}
\label{Stage3}
\textbf{DPO Fine-tuning.} After obtaining the preference dataset $\mathcal{D}_P$, we fine-tune the prover model 
using DPO. 
For each triplet $ \left(s, t_w, t_l \right)$ in $\mathcal{D}_P$, it corresponds to the triplet $ \left(x, y_w, y_l \right)$ in the DPO preference data discussed in \cref{section_3.1}, 
where s corresponds to an input, $t_w$ denotes the preferred response, and $t_l$ denotes the dispreferred response. 
We then minimize the optimization objective of \cref{eq_loss}, where $\pi_{ref}$ corresponds to the prover model before DPO fine-tuning.   

\textbf{Curriculum-based Iteration (CI).} After obtaining the new prover model, the process enters a new round of curriculum-based DPO iteration, as outlined in Algorithm 1. 
Specifically, for the $n$-th iteration, the dataset $\mathcal{C}_n$, with \textit{difficulty} value n, is used as the curriculum data. Tactic sampling is performed using the prover model $\mathcal{P}_{n-1}$ from the previous iteration, and fine-grained scoring is applied to a subset of the data in $\mathcal{C}_n$ to obtain the scoring dataset $\mathcal{SC}_n^0$. This dataset is used to fine-tune the score generator $\mathcal{G}_{n-1}$ from the previous round through supervised learning, yielding the new score generator $\mathcal{G}_n$. Next, $\mathcal{G}_n$ is used to score the remaining data in $\mathcal{C}_n$, resulting in the scoring dataset $\mathcal{SC}_n$. Data filtering and pairing are then performed on $\mathcal{SC}_n$ to obtain the preference pair dataset, followed by a new round of DPO. In the first iteration, $\mathcal{P}_{\text{ref}}$ is the base prover $\mathcal{P}_0$ obtained after SFT. In the $n$-th iteration, $\mathcal{P}_{\text{ref}}$ is $\mathcal{P}_{n-1}$. 

\begin{algorithm}[t]
    \caption{Curriculum-based DPO Iteration}
    \label{alg:dpo_iteration}
    \renewcommand{\algorithmicrequire}{\textbf{Input:}}
    \renewcommand{\algorithmicensure}{\textbf{Output:}}

    \begin{algorithmic}[1]
        \Require {Curriculum dataset $\mathcal{C} = \{ \mathcal{C}_1, \mathcal{C}_2, \dots, \mathcal{C}_I \}$, the base prover model $\mathcal{P}_0$ obtained from SFT, Base score generator model $\mathcal{G}_0$}  
        \Ensure $\mathcal{P}_I$: The final prover model 
        
        \State \textbf{Initialize:}  $\mathcal{P}_\textbf{ref} \gets \mathcal{P}_0$, $\mathcal{P}_\theta \gets \mathcal{P}_\textbf{ref}$
        \For{$n = 1$ \textbf{to} $I$}
            \State Use $\mathcal{C}_n$ as curriculum data.
            \State $\mathcal{T}$ $\gets$ Sample candidate tactics from $\mathcal{C}_n$
            \State $\mathcal{SC}_n^0$ $\gets$ Apply fine-grained scoring to a subset of $\mathcal{C}_n$  to obtain scoring dataset
            \State $\mathcal{G}_n$ $\gets$ SFT on score generator $\mathcal{G}_{n-1}$ with $\mathcal{SC}_0^n$ to obtain new score generator
            \State $\mathcal{SC}_n$ $\gets$ Use $\mathcal{G}_n$ to score the remaining data in $\mathcal{C}_n$ and get scoring dataset
            \State Filter and pair data in $\mathcal{SC}$ to form preference pair dataset $\mathcal{D}_P$
            \State Perform DPO on $\mathcal{P}_{n-1}$:  $\mathcal{P}_{\text{ref}} \gets \mathcal{P}_{n-1}$, $\mathcal{P}_\theta \gets \mathcal{P}_{\text{ref}}$
            \State $\mathcal{P}_n$ $\gets$ Minimize the loss $\mathcal{L}_{DPO}$($\mathcal{P}_\theta$, $\mathcal{P}_{\text{ref}}$) to update $\mathcal{P}_\theta$
        \EndFor
    \State \textbf{return} $\mathcal{P}_n$
    \end{algorithmic}
\end{algorithm} 

\medskip
\section{Experiments}

\subsection{Experimental Setup}
\textbf{Datasets and Metric.} We use the Lean 4 formalized theorem-proving dataset Mathlib4\footnote{https://github.com/leanprover-community/mathlib4} 
as the training set $\mathcal{D}$. 
To evaluate the performance of the proposed method, we used datasets MiniF2F \cite{zheng2021minif2f} and ProofNet \cite{proofnet} as evaluation benchmarks.
MiniF2F consists of 488 problems derived from high school mathematics competitions, with 244 problems in both the test and validation sets. These problems primarily include those sourced from the MATH \cite{hendrycks2021MATH} dataset, high school mathematics competitions such as AMC, AIME, and IMO, as well as a set of carefully crafted problems designed to match the difficulty level of those in the competitions. 
ProofNet includes 371 examples, each comprising a formal theorem statement, a natural language theorem statement, and a natural language proof in Lean 3. 
We manually translated the corresponding Lean 3 proofs in ProofNet into Lean 4, resulting in 360 examples that are suitable for interaction with Lean 4. 
We use pass@k as the evaluation metric, which represents the probability that the model successfully finds at least one valid proof within k proof attempts. 

\textbf{Baselines} We selected different models as the prover model and score generator model to evaluate the effectiveness of the proposed method. Specifically, we chose Mathstral-7B-v0.1\cite{jiang2023mistral7b}, 
Llama3.1-8B-Instruct\cite{dubey2024llama}, 
and Phi-3.5-mini-instruct \cite{abdin2024phi3technicalreporthighly}
as the prover models, and Llama3.1-8B-Instruct, 
gemma-2-9b-it\cite{Riviere2024Gemma2I},  
and gemma-2-2b-it 
as the score generator models for the experiments. We used the entire Mathlib4 dataset for SFT and directly tested the base model, with these two approaches serving as baselines for comparison. Both baseline methods and the proposed method were tested in the Lean 4 
environment to assess their performance.


\textbf{Implementation details} 
The prover’s sampling parameters remain consistent throughout the entire process. We set its temperature to 1.0 and the number of candidate candidate tactics $k$ to 32 for each proof state. In the Preference Data Generation stage, for each curriculum dataset \(C_n\), we randomly select 2000 data points to interact with the Lean environment. After applying the candidate tactic to the current proof state, we obtain a new proof state. Starting from this updated state, we perform 10 search attempts (\(n_{\text{attempt}} = 10\)) and use the frequency of successful proofs as the score of the tactic. For the remaining unscored data, we use the trained score generator \( \mathcal{G}_n \) to evaluate the candidate tactic, setting the temperature to 0.1. For each candidate tactic in each proof state, we perform 10 sampling iterations and compute the average score of these samples as the tactic’s final score. During the DPO fine-tuning process, the hyperparameter \( \beta \) in the loss function is set to 0.1. In the evaluation phase, we limit the search time for each theorem to 300 seconds and employ standard best-first search algorithm to find proofs\cite{NEURIPS2023_44414694, DBLP:journals/corr/abs-2102-06203}. For further details, please refer to Appendix \ref{training details}.

\subsection{Main Results}
To validate the effectiveness of the proposed method, we apply the CuDIP method to three different model groups 
and 
compare the results with existing baseline methods 
as well as the corresponding models after fine-tuning with SFT. 
The proof pass rates on the MiniF2F and ProofNet are presented in 
\cref{tab:main_results}, 
where our method corresponds to the results of the fourth iteration. 
As shown in \cref{tab:main_results}, the use of the CuDIP method leads to significant improvements over the baseline results. For instance, when using Mathstral as the prover and Llama-3.1 as the score generator with CuDIP, the pass@1 on MiniF2F reaches 38.5\%, representing a 4.1\% improvement over the baseline method after SFT. 
Additionally, when using Llama-3.1 as the prover and Gemma2 as the score generator, the highest performance improvements were observed on both benchmarks, with a 7.4\% increase on MiniF2F and a 2.2\% increase on ProofNet. 
In the experiments with the remaining group of models, the proof pass rate also showed notable improvements. 
These results demonstrate the effectiveness of the proposed CuDIP method.

\begin{table*}[htbp]
\caption{
Proving success rates (Pass@1) on the MiniF2F and ProofNet datasets. The superscript \textbf{*} denotes the model as a prover, and \textbf{\dag} indicates it as a score generator.
The highest success rates are \textbf{bolded}. 
}
\label{tab:main_results}
\vskip 0.15in
\begin{center}
\begin{small}
\begin{sc}
\renewcommand{\arraystretch}{1.2} 
\begin{tabular}{l c c c r}
\toprule
\textbf{Method} &  \textbf{MiniF2F-valid} & \textbf{MiniF2F-test} & \textbf{ProofNet}\\ 
\midrule
        Mathstral &23.3\%  &27.0\%  &7.2\%\\ %
        Mathstral + SFT &34.4\% &34.4\% &11.1\%\\ %
        \textbf{Mathstral\textsuperscript{\textbf{*}} + Llama-3.1\textsuperscript{\textbf{\dag}} + CuDIP(ours)} & \textbf{38.5\%} & 
        \textbf{38.5\%} & \textbf{12.7\%} \\ \hline
        
        Llama-3.1&20.1\%  &22.5\% &1.7\%\\ %
        Llama-3.1 + SFT &28.6\% &32.3\% &9.4\%\\ %
        \textbf{Llama-3.1\textsuperscript{\textbf{*}} + gemma2\textsuperscript{\textbf{\dag}} + CuDIP(ours)}  & \textbf{36.0\%} & \textbf{36.8\%} & \textbf{11.6\%} \\ \hline
        
        Phi-3.5 &3.7\% &3.7\% & 0\% \\ %
        Phi-3.5 + SFT  &29.1\%  &32.7\%  & 7.7\%\\ %
        \textbf{Phi-3.5\textsuperscript{\textbf{*}} + gemma2\textsuperscript{\textbf{\dag}} + CuDIP(ours)} 
         & \textbf{33.2\%} & \textbf{36.4\%} & \textbf{9.7\%} \\
        \bottomrule
    \end{tabular}
\end{sc}
\end{small}
\end{center}
\vskip -0.1in
\end{table*}

\begin{figure*}[tbp]
\centering
\includegraphics[width=0.9\linewidth]{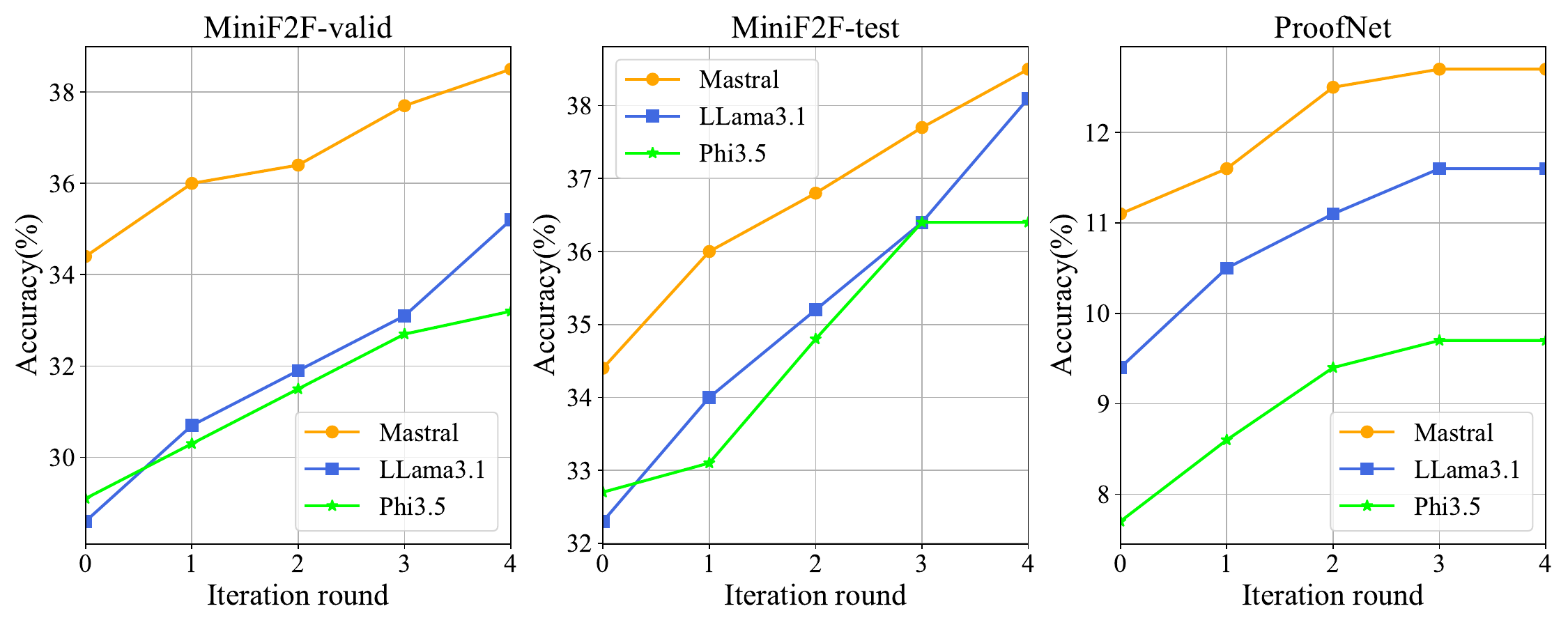}
\caption{Comparison curves of different iteration rounds 
using the CuDIP method. 
The performance curves depicting the proof success rates of the three model groups after four rounds of DPO iteration based on curriculum learning, evaluated on the MiniF2F-valid, MiniF2F-test, and ProofNet benchmarks.
}
\label{iteration}
\end{figure*}

In order to validate the effect of different iteration counts using the CuDIP method, we performed four rounds of DPO iteration on three sets of models. The testing results of the models after each iteration 
are shown in Figure \ref{iteration}. As illustrated in Figure \ref{iteration}, during the first four iterations, the model's proof capabilities increase with the number of iterations, thereby confirming the effectiveness of the proposed iterative method. 
The specific test results of these three groups of models after four iterations can be found in Appendix \ref{appendixA}.

\subsection{Ablation Studies}
To separately investigate the effects of different components of the proposed CuDIP, we conducted ablation experiments, testing: 
1) the effectiveness of the fine-grained preference scoring based on the score generator, 
and 2) the effectiveness of DPO iteration based on curriculum learning.

\textbf{Fine-Grained Preference Scoring (FGPS).}
In order to verify the effectiveness of fine-grained preference scoring during the construction of preference data, we used Mathstral as the prover model and tested two conditions: one where Llama3.1 was used as the score generator in the preference data construction process, and another where no score generator or scoring rules were applied for fine-grained scoring. The test results on the MiniF2F and ProofNet benchmarks are presented in the 
\cref{tab:ablation_result_FGPS}
The use of the score generator for fine-grained scoring significantly improved the model's proof success rate, thereby demonstrating the effectiveness of the proposed fine-grained preference scoring process.

\begin{table*}[t]
\centering
\caption{
Improvement in pass rates on MiniF2F and ProofNet at Pass@1 with 
Fine-grained Preference Scoring (FGPS). 
}
\label{tab:ablation_result_FGPS}
\vskip 0.15in
\begin{center}
\begin{small}
\begin{sc}
\begin{tabular}{l c c c r}
        \toprule
        \textbf{Model} &  \textbf{FGPS}&
        \textbf{MiniF2F-valid} & \textbf{MiniF2F-test} & \textbf{ProofNet}\\ 
    \hline
    Mathstral& \ding{55} &35.2\% &35.6\% &11.6\% \\ 
    Mathstral& \ding{51} &36.0\% &36.0\% &11.6\% \\
    \bottomrule
    \end{tabular}
\vskip -0.1in
\end{sc}
\end{small}
\end{center}
\end{table*}

\textbf{Curriculum-based Iteration (CI).} 
To validate the effect of Curriculum-based Iteration (CI), we combined the curriculum data from the first four iterations and conducted one round of training. The results were then compared with those obtained after four iterations of training. As shown in 
\cref{tab:ablation_result_CL}
, the test results on MiniF2F after training with CI were 4.1\%, showing an improvement of 2.1\% compared to the results without CI, which further verified the effectiveness of curriculum-based iteration.

\begin{table*}[t]
\centering
\caption{Improvement in pass rates for MiniF2F and ProofNet at Pass@1 with 
Curriculum-based Iteration (CI).
}
\label{tab:ablation_result_CL}
\vskip 0.15in
\begin{center}
\begin{small}
\begin{sc}
\begin{tabular}{l c c c r}
        \toprule
        \textbf{Model} &  \textbf{CI}
        & 
        \textbf{MiniF2F-valid} & \textbf{MiniF2F-test} & \textbf{ProofNet}\\ 
    \hline
    Mathstral&  \ding{55} &35.6\% &36.4\% &12.2\%\\ 
   Mathstral(4 iterations) & \ding{51} & 38.5\% & 38.5\% & 12.7\% \\
    \bottomrule
    \end{tabular}
\vskip -0.1in
\end{sc}
\end{small}
\end{center}
\end{table*}

\section{Conclusion}
In this paper, we present a 
curriculum learning-based DPO iterative theorem proving method (CuDIP).
Specifically, we first introduce an effective preference data construction approach based on 
fine-grained preference scoring by 
large language models (LLMs), 
which reduces the reliance on human annotations and enhances the diversity of the preference data. 
Subsequently, we integrate the proposed preference data construction method with curriculum learning to conduct iterative DPO training for the theorem proving model. 
Experimental results demonstrate that the method proposed herein significantly improves the performance of LLMs in theorem proving tasks and enhances their reasoning capabilities in the context of theorem proving.


\bibliographystyle{plain}
\bibliography{reference}

\newpage
\appendix
\onecolumn

\section{More Experimental Results}
\label{appendixA}
In this paper, we used three groups of models to verify our proposed method. For each group of models, we performed 4 rounds of iterations. After each round of iteration, we tested the model. Table 4 shows the test results of the model on the MiniF2F and ProofNet datasets. The results show that the accuracy of the model gradually increases during the iteration process.
\begin{table*}[htbp]
\caption{Proof success rates for Curriculum Learning-based DPO Iterative Theorem Proving (CuDIP) on MiniF2F and ProofNet with Lean at different iterations. Results of our CuDIP method after 4 iterations are shown.}
\label{tab:ablation_result_1}
\vskip 0.15in
\begin{center}
\begin{small}
\begin{sc}
\begin{tabular}{l c c c r}
        \toprule
        \textbf{Model} &  \textbf{Iterations}& 
        \textbf{MiniF2F-valid} & \textbf{MiniF2F-test} 
        & \textbf{ProofNet}\\  
    \hline
    \multirow{4}{*}
        {Mathstral\textsuperscript{\textbf{*}} + Llama-3.1\textsuperscript{\textbf{\dag}} + CuDIP
        } 
        & 1   & 36.0\% & 36.0\% & 11.6\%  \\ 
        & 2   & 36.4\% & 36.8\% & 12.5\%  \\ 
        & 3   & 37.7\% & 37.7\% & 12.7\%  \\ 
        & 4   & 38.5\% & 38.5\% & 12.7\%  \\
        \hline 
        \multirow{4}{*}{Llama-3.1\textsuperscript{\textbf{*}} + gemma2\textsuperscript{\textbf{\dag}} + CuDIP   
        } 
        & 1 &31.9\% &34.0\% &10.8\% \\ 
        & 2 &34.0\% &35.2\% &11.1\% \\ 
        & 3 &35.6\% &36.4\% &11.6\% \\ 
        & 4 &36.0\% &36.8\% &11.6\% \\
        \hline 
        \multirow{4}{*}{Phi-3.5\textsuperscript{\textbf{*}} + gemma2\textsuperscript{\textbf{\dag}} + CuDIP
        } 
        & 1 &30.3\% &33.1\% &8.6\% \\ 
        & 2 &31.5\% &34.8\% &9.4\% \\ 
        & 3 &32.7\% &36.4\% &9.7\% \\ 
        & 4 &33.2\% &36.4\% &9.7\% \\
    \bottomrule
    \end{tabular}
\vskip -0.1in
\end{sc}
\end{small}
\end{center}
\end{table*}

\section{Training Details}
\label{training details}
\subsection{Training Tools}
In this paper, we used supervised fine-tuning (SFT) and direct preference optimization (DPO) to train the model. In terms of training tools, we used LlamaFactory\cite{zheng-etal-2024-llamafactory}, a unified framework that integrates a suite of cutting-edge efficient training methods. It can help us focus more on the construction of the dataset without worrying about the implementation of the model training code.
\subsection{Base Prover Training}
We used LlamaFactory to train the three basic models with SFT. The three models used the same prompt template, see Appendix \ref{prompt_templates} 
for details. During training, we used the LoRA\cite{hu2021lora} method, which can help us save training resources. We set the learning rate to $2.0 \times 10^{-5}$, the learning rate scheduler to cosine, the warmup ratio to 0.03, and trained for 3 epochs. We set the floating point precision to bfloat16 and batch size to 4.
\subsection{Score Generator Training}
When generating training data for the score generator, we performed a Monte Carlo Tree Search (MCTS) on each proof state in the subset of the curriculum learning data $C_n$. For each search, we set the number of simulations to 1000 and limited the search depth to 10. We did not use a fixed value for the search time limit, because as the number of iterations increases, the model's ability gradually increases and the model's search time increases. We observed that when the model performs automatic theorem proving, it usually succeeds within 60 seconds, and the probability of the model proving the theorem after that is very low. This phenomenon is caused by the model's insufficient long-chain reasoning ability, which is also one of the challenges faced by large language models (LLMs) at this stage. In order to balance the quality and quantity of training data when computing resources are limited, we limited the search time limit to 60 seconds in the first round of iterations. For each subsequent round, we increased the search time limit by 30 seconds each time, hoping that the model can perform a deeper search.
\subsection{Filtering and Paring}
After we score each tactic in the candidate tactic set for a proof state, we need to consider how to construct a preferred data set for DPO training. For those tactics with non-zero scores, it means that after the current proof state applies the tactic, the model has the probability of completing the subsequent proof. We use a pair of tactics with a score difference of more than 0.5 as positive and counterexample in a dpo training data set. The advantage of this method is that for any proof tactic, when constructing the DPO training data set, it will not be used as both a positive example and a counterexample, which can be beneficial to the training of the model. When conducting dpo training, we set $\beta$ in the DPO loss function to 0.1, set the learning rate to $5.0 \times 10^{-6}$, and trained for 1 epoch.

\section{Prompt Templates}
\label{prompt_templates}

\subsection{Prove Model Prompt Template}
\begin{tcolorbox}[colframe=brown!40!black, colback=brown!5,title=\textbf{Prompt Template}]
\texttt{<s>You are using Lean4 for theorem proving[INST] You will be provided with a json-formatted data describing the intermediate proof status of the Lean4 theorem. \\ 
You need to give a prediction of a proof tactic based on the data description and your knowledge and experience about Lean4 proofs.\\
The input json format description is as follows: \\ 
\{\\ 
\hspace*{2em}"current\_theorem\_state": <current theorem state description> \\ 
\}\\
Your output must also be json-formatted data in the following format: \\
\{ \\
\hspace*{2em}"predict\_tactic": <A proof tactic that complies with Lean4 syntax. The type is a string and the content cannot contain "sorry"> \\
\} \\
Your input: \\
\{ \\
\hspace*{2em}"current\_theorem\_state": "\$current\_theorem\_state" \\
\} \\
Your output:\\
}
\end{tcolorbox}
We apply this template to an actual proof state and replace the placeholders in the template about the theorem proof state with specific values. Considering that the proof state of a theorem may contain multiple proof goals, we set the upper limit of the number of model input tokens to 4096 and the upper limit of the number of output tokens to 2048, which can meet the needs of the current task. It is worth mentioning that in order to better help us extract the desired tactics from the model output, we limit the input and output of the model to json format. \\
The following is a actual examples of prompts and the responses of the model:

\begin{tcolorbox}[colframe=brown!40!black, colback=brown!5, breakable, title=\textbf{Input Example}]
\texttt{<s>You are using Lean4 for theorem proving[INST] You will be provided with a json-formatted data describing the intermediate proof status of the Lean4 theorem. \\ 
You need to give a prediction of a proof tactic based on the data description and your knowledge and experience about Lean4 proofs.\\
The input json format description is as follows: \\ 
\{\\ 
\hspace*{2em}"current\_theorem\_state": <current theorem state description> \\ 
\}\\
Your output must also be json-formatted data in the following format: \\
\{ \\
\hspace*{2em}"predict\_tactic": <A proof tactic that complies with Lean4 syntax. The type is a string and the content cannot contain "sorry"> \\
\} \\
Your input: \\
\{ \\
\hspace*{2em}"current\_theorem\_state": "a b c : $\mathbb{N}$\texttt{\textbackslash n} $h_1$ : b = 0\texttt{\textbackslash n} $\mathbb{\vdash}$ a + b + c = c + a" \\
\} \\
Your output:\\
}
\end{tcolorbox}

For the above input, LLM's response is as follows
\begin{tcolorbox}[colframe=brown!40!black, colback=brown!5, title=\textbf{Output Example}]
\texttt{\{ \\
\hspace*{2em}"predict\_tactic": "norm\_num [$h_1$]" \\
\} \\
}
\end{tcolorbox}

\subsection{Score Generator Prompt Template}
\begin{tcolorbox}[colframe=brown!40!black, colback=brown!5, breakable,title=\textbf{Prompt Template}]
\texttt{
\#\#\# Task Description: \\
You will be given a `tactic` generated by a model in the context of a Lean4 theorem proving process. Please score the `tactic` based on the following criteria, where: \\
- 0 means the `tactic` cannot run in Lean4 syntax or does not affect the current proof state.\\
- 1 means the `tactic` fully complies with Lean4 syntax and successfully completes the proof.\\
- A value between 0 and 1 indicates the probability of the `tactic` completing the proof or significantly advancing the proof process.\\
Do not provide any additional text, explanations, or responses. Only output the score which is a number between 0 and 1.\\
\\
\#\#\# Output Format: \\
- A score (floating-point value): A number between 0 and 1.\\
\\
\#\#\#\# Input: \\
\{\\
\hspace*{2em}"current\_theorem\_state": "\$current\_theorem\_state"\\
\}\\
\\
\#\#\#\# Generated `tactic`: \\
\{\\
\hspace*{2em}"predict\_tactic": "\$predict\_tactic"\\
\}\\
\\
\texttt{---}\\
\\
\#\#\# Score:\\
}
\end{tcolorbox}

\section{Case Studies}
\subsection{mathd\_algebra\_109}
Theorem mathd\_algebra\_109 in the valid validation set of Minif2f is defined as follows:
\begin{tcolorbox}[colframe=brown!40!black, colback=brown!5,title=\textbf{Theorem Define}]
theorem mathd\_algebra\_109 (a b : $\mathbb{R}$) ($h_0$ : 3 * a + 2 * b = 12) ($h_1$ : a = 4) : b = 0
\end{tcolorbox}

Each time we replace the theorem state placeholder in the prompt with the intermediate state when proving the theorem, and then use the proof tactic given by the model to prove the theorem.

\begin{tcolorbox}[colframe=brown!40!black, colback=brown!5, breakable,title=\textbf{Example}]
\texttt{<s>You are using Lean4 for theorem proving[INST] You will be provided with a json-formatted data describing the intermediate proof status of the Lean4 theorem. \\ 
You need to give a prediction of a proof tactic based on the data description and your knowledge and experience about Lean4 proofs.\\
The input json format description is as follows: \\ 
\{\\ 
\hspace*{2em}"current\_theorem\_state": <current theorem state description> \\ 
\}\\
Your output must also be json-formatted data in the following format: \\
\{ \\
\hspace*{2em}"predict\_tactic": <A proof tactic that complies with Lean4 syntax. The type is a string and the content cannot contain "sorry"> \\
\} \\
Your input: \\
\{ \\
\hspace*{2em}"current\_theorem\_state": "b : $\mathbb{R}$\texttt{\textbackslash n} $h_0$ : 3 * 4 + 2 * b = 12\texttt{\textbackslash n} b = 0" \\
\} \\
Your output:\\
}
\tikz \draw[dashed] (0,0) -- (\linewidth,0);
\texttt{
\{ \\
\hspace*{2em}"predict\_tactic": " linarith" \\
\} \\
}
\end{tcolorbox}

Theorem mathd\_algebra\_109 can be proved using two proof tactics, \texttt{"}subst $h_1$\texttt{"} and \texttt{"}linarith\texttt{"}, but this does not mean that there is only one way to prove the theorem. When the model predicts the tactics for the first proof state, it will not only generate the tactic shown above, but also other proof tactics. If the proof tactic of the first step, \texttt{"}subst $h_1$\texttt{"}, is replaced by \texttt{"}rw [$h_1$] at $h_0$\textbf{\texttt{"}}, and the proof tactic of the second step is not changed, the theorem can also be proved.

\subsection{mathd\_numbertheory\_109}
The theorem is defined as follows
\begin{tcolorbox}[colframe=brown!40!black, colback=brown!5,title=\textbf{Theorem Define}]
theorem mathd\_algebra\_109 (v : $\mathbb{N}$ → $\mathbb{N}$) ($h_0$ : $\forall$ n, v n = 2 * n - 1) :
  ($\sum$ k in Finset.Icc 1 100, v k) \% 7 = 4
\end{tcolorbox}

\begin{tcolorbox}[colframe=brown!40!black, colback=brown!5, breakable,title=\textbf{Step 1}]
\texttt{<s>You are using Lean4 for theorem proving[INST] You will be provided with a json-formatted data describing the intermediate proof status of the Lean4 theorem. \\ 
You need to give a prediction of a proof tactic based on the data description and your knowledge and experience about Lean4 proofs.\\
The input json format description is as follows: \\ 
\{\\ 
\hspace*{2em}"current\_theorem\_state": <current theorem state description> \\ 
\}\\
Your output must also be json-formatted data in the following format: \\
\{ \\
\hspace*{2em}"predict\_tactic": <A proof tactic that complies with Lean4 syntax. The type is a string and the content cannot contain "sorry"> \\
\} \\
Your input: \\
\{ \\
\hspace*{2em}"current\_theorem\_state": "v : $\mathbb{N}$ → $\mathbb{N}$ $h_0$ : $\forall$ (n : $\mathbb{N}$), v n = 2 * n - 1 $\mathbb{\vdash}$ ($\sum$ k $\in$ Finset.Icc 1 100, v k) \% 7 = 4" \\
\} \\
Your output:\\
}
\tikz \draw[dashed] (0,0) -- (\linewidth,0);
\texttt{
\{ \\
\hspace*{2em}"predict\_tactic": "norm\_num $h_0$" \\
\} \\
}
\end{tcolorbox}

\begin{tcolorbox}[colframe=brown!40!black, colback=brown!5, breakable,title=\textbf{Step 2}]
\texttt{<s>You are using Lean4 for theorem proving[INST] You will be provided with a json-formatted data describing the intermediate proof status of the Lean4 theorem. \\ 
You need to give a prediction of a proof tactic based on the data description and your knowledge and experience about Lean4 proofs.\\
The input json format description is as follows: \\ 
\{\\ 
\hspace*{2em}"current\_theorem\_state": <current theorem state description> \\ 
\}\\
Your output must also be json-formatted data in the following format: \\
\{ \\
\hspace*{2em}"predict\_tactic": <A proof tactic that complies with Lean4 syntax. The type is a string and the content cannot contain "sorry"> \\
\} \\
Your input: \\
\{ \\
\hspace*{2em}"current\_theorem\_state": "v : $\mathbb{N}$ → $\mathbb{N}$ $h_0$ :$\forall$ (n : $\mathbb{N}$), v n = 2 * n - 1 $\mathbb{\vdash}$ ($\sum$ x $\in$ Finset.Icc 1 100, 2 * x - 1) \% 7 = 4" \\
\} \\
Your output:\\
}
\tikz \draw[dashed] (0,0) -- (\linewidth,0);
\texttt{
\{ \\
\hspace*{2em}"predict\_tactic": "rfl" \\
\} \\
}
\end{tcolorbox}



\end{document}